\documentclass[10pt,twocolumn,letterpaper]{article}
\usepackage{times}
\usepackage{longtable}
\usepackage{caption}
\usepackage{latexsym}
\usepackage{listings}
\usepackage{rotating}
\usepackage{cvpr}      
\usepackage[T1]{fontenc}

\usepackage{microtype}
\usepackage{inconsolata}
\colorlet{punct}{red!60!black}
\definecolor{background}{HTML}{FFFFFF}
\definecolor{delim}{RGB}{20,105,176}
\colorlet{numb}{magenta!60!black}

\lstdefinelanguage{json}{
    basicstyle=\normalfont\ttfamily,
    numbers=left,
    numberstyle=\scriptsize,
    stepnumber=1,
    numbersep=8pt,
    showstringspaces=false,
    breaklines=true,
    frame=none,
    backgroundcolor=\color{background},
    literate=
     *{0}{{{\color{numb}0}}}{1}
      {1}{{{\color{numb}1}}}{1}
      {2}{{{\color{numb}2}}}{1}
      {3}{{{\color{numb}3}}}{1}
      {4}{{{\color{numb}4}}}{1}
      {5}{{{\color{numb}5}}}{1}
      {6}{{{\color{numb}6}}}{1}
      {7}{{{\color{numb}7}}}{1}
      {8}{{{\color{numb}8}}}{1}
      {9}{{{\color{numb}9}}}{1}
      {:}{{{\color{punct}{:}}}}{1}
      {,}{{{\color{punct}{,}}}}{1}
      {\{}{{{\color{delim}{\{}}}}{1}
      {\}}{{{\color{delim}{\}}}}}{1}
      {[}{{{\color{delim}{[}}}}{1}
      {]}{{{\color{delim}{]}}}}{1},
}
\usepackage[utf8]{inputenc}
\usepackage{graphicx}
\usepackage{amsmath}
\usepackage{booktabs} 
\usepackage{multirow}
\usepackage{tabularx}
\usepackage[most]{tcolorbox}
\usepackage[table,usenames,dvipsnames]{xcolor}
\usepackage{algorithm,algpseudocode}

\definecolor{cvprblue}{rgb}{0.21,0.49,0.74}
\usepackage[pagebackref,breaklinks,colorlinks,allcolors=cvprblue]{hyperref}

\def\paperID{*****} 
\def\confName{CVPR}
\def\confYear{2026}

\title{Role-SynthCLIP: A Role Play Driven Diverse Synthetic Data Approach}

\author{
Yuanxiang Huangfu\textsuperscript{1} \quad
Chaochao Wang\textsuperscript{1} \quad
Weilei Wang\textsuperscript{1}  \\
\textsuperscript{1}PatSnap Co., LTD. \\
{\tt\small \{huangfuyuanxiang, wangchaochao, wangweilei\}@patsnap.com} \\
}

\begin{document}
\maketitle
\begin{abstract}
The effectiveness of Contrastive Language-Image Pre-training (CLIP) models critically depends on the semantic diversity and quality of their training data. However, while existing synthetic data generation methods primarily focus on increasing data volume, such emphasis often leads to limited semantic diversity and redundant or shallow captions. To address this limitation, we propose \textbf{Role-SynthCLIP}, a novel data synthesis framework that leverages multi-perspective role-playing prompts (e.g., a compositional analyst, an interpreter of image context) to guide Multimodal Large Language Models (MLLMs) in generating semantically diverse captions from distinct viewpoints. This mechanism enhances the semantic diversity and fine-grained image-text alignment of synthetic pairs, thereby improving caption expressiveness and accuracy while keeping the total number of image-text pairs unchanged. Experimental results demonstrate the effectiveness and efficiency of our method. A CLIP-B/16 model trained on only 1 million Role-SynthCLIP pairs achieves a Recall@1 of $\mathbf{64.1\%}$ on the MS COCO validation set, surpassing the best existing synthetic data baseline (trained on 5M pairs) by $2.8$ percentage points. The code and trained models are released at \url{https://github.com/huangfu170/Role-SynthCLIP}.
\end{abstract}    
\section{Introduction}
The paradigm of vision-language models (VLMs), such as CLIP \cite{clip}, has profoundly reshaped multimodal learning by aligning vision and language representations \cite{luo2023lexliplexiconbottleneckedlanguageimagepretraining}. Traditionally, their success has relied on massive web-crawled datasets containing millions to billions of image-text pairs \cite{zhang2024magiclensselfsupervisedimageretrieval}. However, this pursuit of data scale—from millions to billions of pairs—has created a critical bottleneck: the growing data volume demands prohibitive computational resources and inevitably incorporates noisy, low-quality, and redundant web samples \cite{eva-clip,eva-clipv2}.

In this context, controllable high-quality synthetic data generation has emerged as a promising alternative to large-scale web scraping. Methods such as SynthCLIP \cite{hammoud2024synthclipreadyfullysynthetic} have demonstrated that competitive performance can be achieved with fully synthetic corpora of up to 30M pairs. Nevertheless, as attention shifts from data scale to data quality, a new challenge arises: semantic impoverishment. Captions generated by generic prompts often exhibit limited diversity: they tend to be uniform, single-perspective, and semantically repetitive, failing to capture the rich, multifaceted nature of visual content. This redundancy constrains the diversity and discriminative capacity of learned visual–language representations.

\begin{figure}
    \centering
    \includegraphics[width=\linewidth]{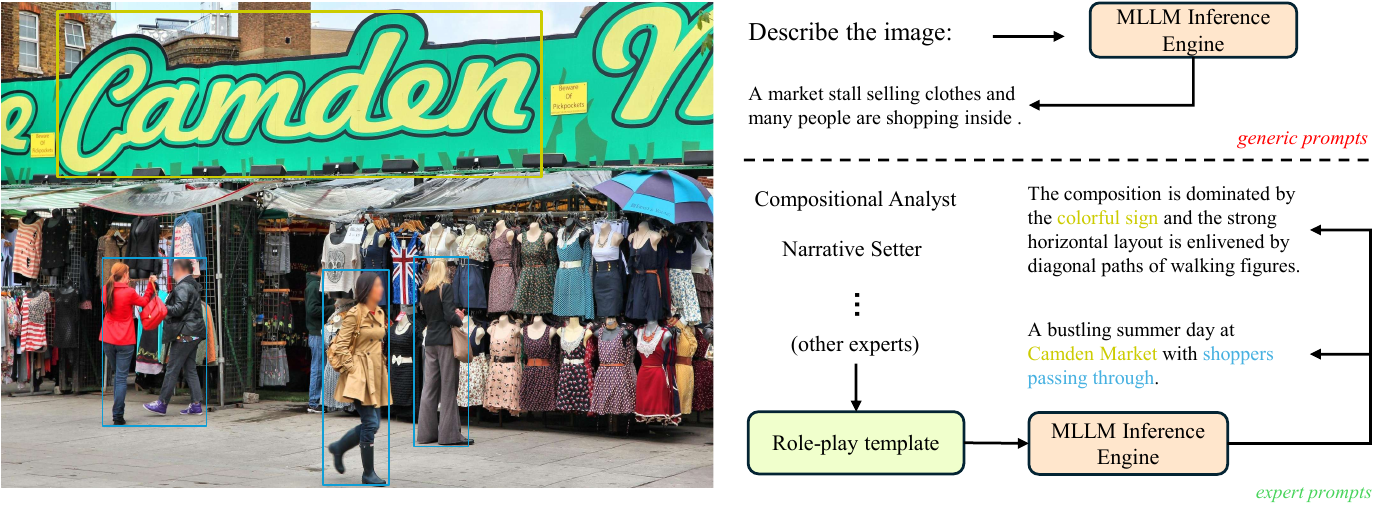}
    \caption{Comparison of synthetic data generation paradigms. Our proposed Role-SynthCLIP framework addresses the semantic impoverishment of existing methods by leveraging multi-expert role-playing prompts. \textbf{(Top)} Under the conventional paradigm, generic prompts fed into the MLLM yield shallow, single-perspective descriptions. \textbf{(Bottom)} Our approach introduces role-play templates (e.g., Compositional Analyst, Narrative Setter), which guide the MLLM to perceive the image from diverse cognitive perspectives. This mechanism generates semantically rich captions that focus on fine-grained elements (e.g., specific objects, visual structure, context), maximizing the diversity of representations learned during VLM pre-training.}
    \label{fig:1}
\end{figure}

Existing efforts to improve synthetic data quality primarily target distributional balance or prompt variety. For instance, SynthCLIP \cite{hammoud2024synthclipreadyfullysynthetic} maintains entity distribution balance across the corpus. Other concurrent works, like FIX-CLIP \cite{wang2025fixclipdualbranchhierarchicalcontrastive}, attempt to capture diversity by employing ensembles of multiple generic prompts (e.g., 20 variants). However, these ensemble approaches lack cognitive depth: their variations are often shallow paraphrases that do not introduce distinct interpretative dimensions-such as the compositional, contextual, or emotional perspectives-of the same image.

To address this limitation, we propose \textbf{Role-SynthCLIP}, a novel framework that amplifies caption diversity by simulating multi-expert perspectives through role-playing prompts. As illustrated in Figure \ref{fig:1}, our method assigns distinct expert roles to a Multimodal Large Language Model (MLLM). For instance, an \textit{Observer of Details} highlights fine-grained attributes, while a \textit{Compositional Analyst} focuses on global structure and visual organization. By combining these complementary perspectives, Role-SynthCLIP produces semantically rich and well-aligned captions that better reflect the complexity of visual scenes.

Our design is motivated by the proven effectiveness of role-playing prompting in LLMs, which has been shown to elicit diverse responses and improve task performance without architectural modifications or additional training \cite{zhao2025roleplayparadoxlargelanguage,kong-etal-2024-better}. By extending this paradigm to MLLMs for caption generation, we can efficiently and flexibly steer the model to produce the desired semantic diversity.

In summary, the main contributions of this paper can be
summarized as threefold:
\begin{itemize}
    \item We propose Role-SynthCLIP, a novel data synthesis framework. By leveraging role-playing prompts, it enables an MLLM to perceive an image from diverse, multi-faceted perspectives (e.g., compositional, detailed, contextual), generating captions with superior semantic diversity and alignment.
    \item We establish new state-of-the-art (SOTA) performance for data-efficient CLIP training. A CLIP-B/16 model trained on only 1M Role-SynthCLIP pairs attains a Recall@1 of 64.1\% on COCO-val, surpassing the best 5M-pair synthetic baseline.
    \item Our work includes a detailed ablation study and empirical analysis across 8 downstream tasks, providing deep insights into the crucial impact of various expert role combinations on the learned representation diversity and model performance.
\end{itemize}
\section{Related Works}

\subsection{Vision-Language Models} 
Recent years have witnessed rapid progress in vision-language models (VLMs) that bridge visual perception and natural language understanding. Among them, Contrastive Language-Image Pretraining (CLIP) \cite{clip,CLIC,chen-etal-2023-altclip,lai2024veclipimprovingcliptraining} has emerged as a representative framework, demonstrating how large-scale image-text alignment can lead to transferable multimodal representations. 

Subsequent studies have investigated extensions of this paradigm, including scaling pretraining corpora \cite{eva-clip,eva-clipv2}, improving robustness under distribution shifts \cite{nguyen2023qualityquantityinteractiondataset}, and adapting CLIP-style embeddings for tasks that require multimodal reasoning tasks \cite{guzhov2021audioclipextendingclipimage,alyami2025clipslaparameterefficientclipadaptation,Zhao_2025}. Despite this progress, a key challenge remains in effectively increasing semantic richness and alignment of the training data, particularly when operating within a constrained training corpus or relying on synthetic generation. Our work specifically addresses this quality bottleneck in the constrained data budget.

\subsection{Synthetic Captions for VLM training} 
Recent advances have demonstrated that caption diversity is paramount for improving the quality of contrastive pre-training. Prior work has explored data volume expansion \cite{xu2024demystifyingclipdata,chuang2025metaclip2worldwide}, simple caption rewriting \cite{liu2024clipsenhancedclipframework}, or linguistic transformations. More recently, with the development of large multimodal models, it has become possible to generate relevant captions for images \cite{lai2024veclipimprovingcliptraining,zheng2024dreamliplanguageimagepretraininglong}. While works like SynthCLIP \cite{hammoud2024synthclipreadyfullysynthetic} focus on balancing entity distributions, and others like FIX-CLIP \cite{wang2025fixclipdualbranchhierarchicalcontrastive} use multiple fixed prompts for variations. However, these generative approaches are typically constrained by generic prompts. They produce repetitive and shallow descriptions from a narrow viewpoint, resulting in surface-level paraphrasing rather than deep cognitive diversity. 

In contrast, our proposed Role-SynthCLIP systematically guides the MLLM through multi-experts role-playing prompts, enabling it to describe an image from diverse semantic perspectives. This approach unlocks the latent reasoning capacity of MLLMs to produce captions that are both fine-grained and semantically complementary.

\subsection{Role-Play Prompting}
Role-playing prompting is a powerful, training-free technique widely employed in Large Language Models (LLMs) across specialized domains, such as education \cite{dan2023educhatlargescalelanguagemodelbased}, healthcare \cite{abbasian2024conversationalhealthagentspersonalized,yu2025healthllmpersonalizedretrievalaugmenteddisease}, and programming \cite{qian2024chatdevcommunicativeagentssoftware}. This mechanism explicitly defines a role's expertise, enabling the model to instantiate a specific persona and activate its inherent, specialized knowledge, leading to substantially enhanced performance on target tasks \cite{kong-etal-2024-better}. While highly successful in text-only domains, the structured use of multi-role prompting for cross-modal data synthesis, particularly for enriching image–text pairs, remains largely unexplored. To the best of our knowledge, Role-SynthCLIP is the first framework to systematically extend multi-role prompting to multimodal large language models for generating semantically diverse and high-quality visual captions.
\section{Methodology}
\begin{figure*}[htbp]
    \centering
    \includegraphics[width=\linewidth]{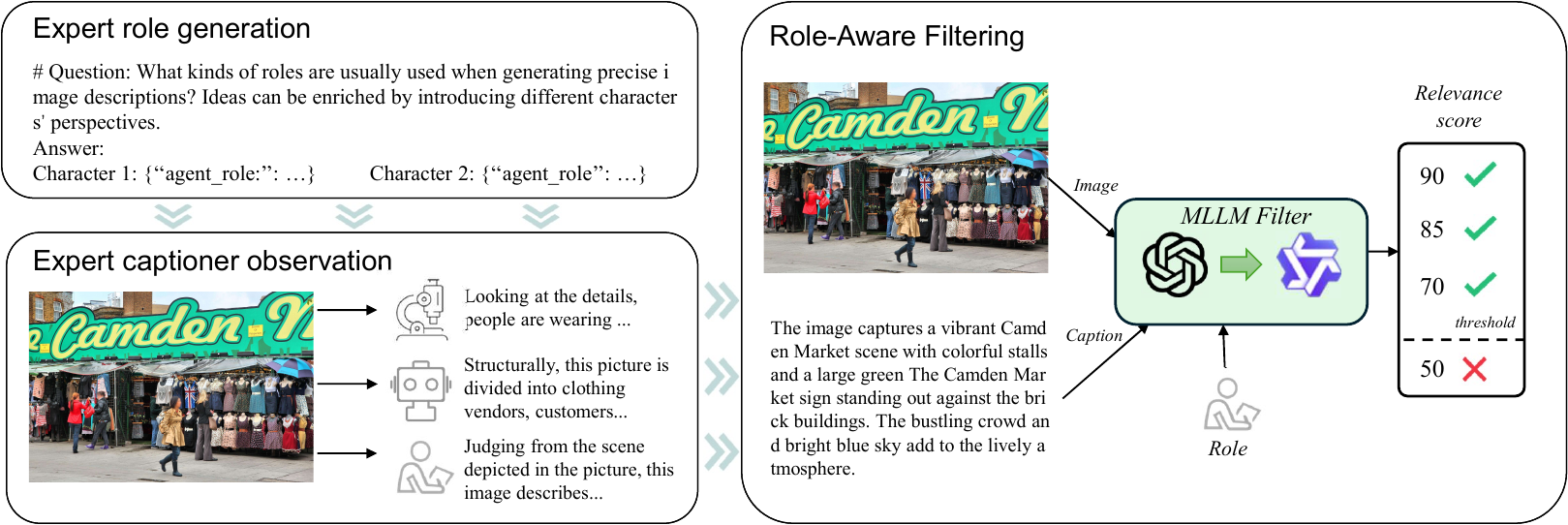}
    \caption{Overview of our Role-SynthCLIP framework. We begin by sampling multiple distinct expert roles from a large language model, each with a defined specialty and set of responsibilities. Subsequently, these expert annotators describe an image from their specific professional viewpoint. Finally, we employ a filtering model to remove potential noise from the generated data, ensuring that the descriptions are both accurate to the image content and consistent with the assigned roles.}
    \label{fig:2}
\end{figure*}
We formulate the task as synthesizing semantically diverse and accurate image-text pairs for contrastive training, given an image corpus $\mathcal{I} = \{I_i\}$ without paired captions. As depicted in Figure~\ref{fig:2}, the Role-SynthCLIP framework operates via a three-stage pipeline designed to synthesize semantically diverse and high-fidelity image-text pairs. The process begins with expert role generation to define specialized annotation personas. This is followed by expert captioner observation, where a multimodal large language model generates multi-perspective captions by strictly adhering to each role's persona. Finally, a dedicated Role-Aware Filter rigorously validates and prunes the synthesized data to ensure both visual accuracy and role-consistency.

\subsection{Expert Role Generation}

A single image can be interpreted through many perspectives. However, instead of using randomly selected prompts, we argue that each prompt should be intentionally designed to focus on a specific aspect of capability, so that they can complement each other as a whole. Inspired by structured agent generation in LLM-Discussion \cite{zhao2025roleplayparadoxlargelanguage}, we design a two-stage process for constructing high-quality expert roles.

In the first stage, we prompt a LLM to propose an initial set of roles suitable for generating precise and informative image descriptions. Subsequently, a two-round conversational process is utilized to convert this high-level information into a structured format, which includes the expert's name, their specific specialty, and their corresponding responsibilities. This two-stage approach ensures the roles are both specialized and syntactically consistent.  These roles are strategically chosen to span critical dimensions of image understanding: fine-grained features (Observer), scene context (Interpreter), visual structure (Compositional Analyst), implied narrative (Narrative Setter), and subjective feeling (Emotional Responder). The specific prompts utilized are detailed in Appendix \ref{sec:appendix_expert_role_prompts}.

\subsection{Expert Captioner Observation}

 We utilize Qwen2.5 VL-7B \cite{bai2025qwen25vltechnicalreport} as our core multimodal captioner, chosen for its high visual grounding and instruction adherence capacity, to generate captions across varying granularities (short and long). To maximize semantic density, the MLLM strictly adheres to each role's persona when generating multi-perspective captions. The specific prompts used for this process are provided in Appendix \ref{sec:appendix_expert_captioner_prompts}. 
 
 To maintain data quality and prevent the introduction of generic or title-like noise, we employ a simple yet effective pre-filtering strategy based on caption length. Long captions with fewer than 10 words and short captions with fewer than 4 words are discarded. Through these straightforward yet effective methods, we obtained a multi-perspective synthetic dataset, in which the average word count of long captions is 92.4 and that of short captions is 17.7.

\subsection{Role-Aware Filtering}
 Generated captions from MLLMs are inherently susceptible to visual inaccuracies, including category and attribute hallucination \cite{li-etal-2023-evaluating, bai2025hallucinationmultimodallargelanguage}. Furthermore, Our multi-expert strategy further requires ensuring consistency with the assigned role’s perspective. To address this, we design a Role-Aware Filter.

Inspired by prior work on finetuned MLLM as validator \cite{wang2024finetunedmultimodallanguagemodels}, we design a more rigorous and interpretable filtering mechanism for our Role-SynthCLIP. To instill highly reliable and interpretable judgment capacity into our Role-Aware Filter, we employ a sophisticated knowledge distillation strategy. We leverage a large, proprietary model, GPT-5, as the teacher, and fine-tune the lightweight Qwen2.5 VL-3B as the student model. The goal is to transfer the teacher's nuanced ability to evaluate both visual fidelity and role-consistency. The training process involves two phases:

\noindent \textbf{Teacher Data Generation.} We sample a diverse set of generated image-text pairs ($I, C, R$) and prompt the GPT-5 teacher, conditioned on the image, the caption, and the expert role's responsibilities. The teacher is instructed to output two correlated responses: a relevance score $S_{\text{GPT-5}}$ (e.g., a 1-100 continuous value) and a detailed rationale $T_{\text{rationale}}$ explaining the judgment based on the three inputs.

\noindent \textbf{Multi-Task Knowledge Distillation.} The Qwen2.5 VL-3B student is then fine-tuned via a multi-task objective to simultaneously mimic the teacher's score and generate the corresponding rationale. This dual-task training extends beyond standard scoring-based filtering, ensuring the student learns what to filter and why.

Our filter lies in its multi-input structure (three-way conditioning): the validator is simultaneously conditioned on the Image ($I$), the generated Caption ($C$), and the expert role's specific Responsibilities ($R$). This allows the filter to score not only visual accuracy but, critically, the semantic consistency with the required perspective. 

The relevance score is formally quantified as:
$$\text{Relevance Score} = MLLM_{filter} (I, C, R)$$

The MLLM filter takes the concatenation of the three inputs ($I$, $C$, and $R$) to produce a final relevance score, ensuring rigorous validation against the specific prompt constraints. Based on these scores, a fixed percentage of image–text pairs with the lowest relevance are filtered out to enhance overall dataset quality. The design of the filter prompt is presented in Appendix \ref{sec:appendix_filter_prompt}, while The detailed pseudocode describing the filtering process is provided in Appendix \ref{sec:appendix_filter_algorithm}.

\subsection{Training Objective and Sequence Length Adaptation}
Standard CLIP model was trained on relatively short captions, with its text encoder supporting a maximum sequence length of only 77 tokens. Given that our synthetic long captions average over 90 words, we adopt the positional embedding extension strategy from Long-CLIP \cite{longclip} to fully utilize the richer semantic content. This approach involves freezing the original positional embeddings ${PE}_{\text{orig}} \in \mathbb{R}^{n \times d}$ for the first 20 tokens and extending the embeddings for subsequent tokens via linear interpolation, based on the observation that initial tokens carry disproportionately critical information. The extension of the positional embedding ${PE}_{\text{long}} \in \mathbb{R}^{m \times d}$ is formally defined as:

$$PE_{\text{long}} = \text{Concat}(PE_{\text{orig}}[: 20], \text{Intpol}(PE_{\text{orig}}[20:], q))$$
where $\mathrm{Intpol}$ denotes the linear interpolation function:
\begin{align*}
\operatorname{Intpol}(PE,q)[i]
  &= (1-\lambda)\,PE_{\mathrm{orig}}[j] + \lambda\,PE_{\mathrm{orig}}[j+1],\\
\lambda &= \tfrac{i\bmod q}{q},\qquad j=\big\lfloor i/q \big\rfloor .
\end{align*}


\noindent where $\operatorname{Intpol}$ denotes the linear interpolation function, the symbol $\lfloor \cdot \rfloor$ denotes the floor function. The variables $q$ and $i$ respectively represent the index of the interpolation ratio and the interpolation position, while $\lambda$ denotes a specified weight.

By this means, the CLIP model is capable of supporting text inputs of up to 248 tokens, effectively allowing it to capture features from our high-density, long-form text. To further exploit the multi-caption characteristics of our synthesized dataset, we modify the standard CLIP contrastive objective into a multi-positive variant.
Instead of assigning a single one-hot label for each image–text pair, we treat all captions belonging to the same image as valid positive samples.
Formally, given a batch of $B$ image–text pairs and an image–caption correspondence matrix $M \in \{0,1\}^{B\times B}$ where $M_{ij}=1$ if image $i$ and text $j$ originate from the same image, the image-to-text loss is defined as:
\begin{equation}
    \mathcal{L}_{i2t}= -\frac{1}{B}\sum_i \sum_j \frac{M_{ij}}{\sum_k M_{ik}}\log \frac{\exp{(s_{ij}/\tau)}}{\sum_l \exp(s_{il}/\tau)}
\end{equation}

\noindent and the text-to-image loss $\mathcal{L}_{t2i}$ is computed symmetrically.
The final training objective is $\mathcal{L}=\frac{1}{2}(\mathcal{L}_{i2t}+\mathcal{L}_{t2i})$.
This variant encourages all captions of the same image to be simultaneously aligned with the corresponding visual embedding, avoiding the false-negative problem inherent to one-hot contrastive objectives when multiple semantically valid captions exist.

From a probabilistic standpoint, the necessity of this modification can be justified by analyzing the expected batch collision rate under our dataset scale.
Given approximately 1M unique images and a global batch size of 2048, the probability that at least two captions from the same image appear within the same batch exceeds about 80\% (following a birthday-paradox approximation \cite{math12243882}).
Therefore, the one-hot objective would incorrectly penalize genuine positive pairs as negatives in the majority of batches, leading to representation inconsistency and reduced training stability.
By adopting the multi-positive formulation, we explicitly correct this bias, ensuring that all semantically equivalent captions reinforce—rather than compete with—each other during contrastive optimization.


\section{Experiments}
\subsection{Experimental Setup}
\noindent \textbf{Datasets.}
To comprehensively evaluate Role-SynthCLIP, we utilize ShareGPT4V \cite{chen2023sharegpt4vimprovinglargemultimodal} as our training dataset, which contains 1M images and is used in Long-CLIP \cite{longclip} and FIX-CLIP \cite{wang2025fixclipdualbranchhierarchicalcontrastive}. We conduct extensive evaluations on widely-used benchmarks, including both zero-shot retrieval and classification. 

For the retrieval task, we evaluate our method on MS-COCO \cite{chen2015microsoftcococaptionsdata}, Flickr-1k \cite{flickr}, Urban-1k \cite{longclip} and report the Recall@1, which is selected to assess the model’s ability to align image and text representations under diverse domains. For classification benchmarks, we utilize commonly used datasets ImageNet-1k \cite{imagenet15russakovsky},  ImageNet-O \cite{srivastava2022distributiondetectionimageneto}, CIFAR-10 \cite{Krizhevsky09learningmultiple}, CIFAR-100 \cite{Krizhevsky09learningmultiple} and Food-101 \cite{bossard14} and report the top-1 accuracy (Acc@1), which measures generalization and robustness, particularly under out-of-distribution (OOD) shifts.

\noindent \textbf{Baseline Models.} We select a variety of models as baselines for comparison, including OpenAI CLIP-ViT-base-patch16 \cite{clip} and CLIP-ViT-large-patch14, the context-extended method Long-CLIP \cite{longclip} with positional extension, and the structurally improved method FIX-CLIP \cite{wang2025fixclipdualbranchhierarchicalcontrastive} built upon it. 

FIX-CLIP is evaluated under different training scales. To demonstrate the effectiveness of our method, we choose its experimental results on 1M and 5M datasets for comparison. We also include SynthCLIP \cite{hammoud2024synthclipreadyfullysynthetic}, which trains CLIP entirely on synthetic image-text pairs generated via LLM-guided text-to-image synthesis, serving as a strong fully-synthetic baseline. 

Finally, we further consider approaches with alternative training objectives, such as SigLIP \cite{zhai2023sigmoidlosslanguageimage} and LoTLIP \cite{wu2024lotlipimprovinglanguageimagepretraining}, ensuring a fair and diverse comparison against state-of-the-art methods.

\noindent \textbf{Training setup.}
To ensure a fair comparison, our experimental setup is consistent with the methodology of Long-CLIP \cite{longclip}. Unless otherwise specified, our models are trained on the ShareGPT4V \cite{chen2023sharegpt4vimprovinglargemultimodal} dataset, which consists of 1M images. 

We employ two canonical variants of CLIP in our experiments, which represent different model scales and allow us to assess robustness across capacities, both of which use a standard Transformer as the text encoder. The input images are resized to 224$\times$224. Following the methodology of Long-CLIP \cite{longclip}, captions are truncated or padded to 248 tokens. Training is conducted on 8 A800 GPUs with Accelerate and Deepspeed \cite{rajbhandari2020zeromemoryoptimizationstraining}. All hyperparameters (learning rate, warmup steps, weight decay, training epochs) are kept consistent with Long-CLIP \cite{longclip} for comparability.

\subsection{Experimental Results}
In this section, we present a detailed evaluation of the Role-SynthCLIP framework on several public benchmarks, including zero-shot multimodal retrieval and classification. We conduct a detailed comparison against state-of-the-art methods to demonstrate the effectiveness and efficiency of our synthetic data approach.

\noindent \textbf{Zero-shot Retrieval.} 
Our results demonstrate that Role-SynthCLIP establishes a new state-of-the-art in low-resource CLIP training. For the CLIP-B/16 architecture, Role-SynthCLIP achieves an average Recall@1 of 77.01\% across all six metrics. This represents a substantial improvement over methods trained on comparable 1M data volumes, such as Long-CLIP (+7.48\% absolute gain), and significantly surpasses the performance of models trained on vastly larger, less curated corpora, including FIX-CLIP (5M, 75.95\%), SigLIP (12B), and LoTLIP (100M). This empirically validates our core hypothesis: quality, curated diversity effectively overcomes the reliance on data volume.

The trend is amplified in the CLIP-L/14 setting, where Role-SynthCLIP achieves an average Recall@1 of 80.43\%, outperforming the best comparable 1M baseline (Long-CLIP) by 5.85\%.

We note that the improvement in Text-to-Image (T2I) retrieval is more moderate, particularly on COCO (43.2\% R@1 for B/16), falling behind the 5M FIX-CLIP result. This observation points to a semantic distillation bottleneck in the text encoder. Although Role-SynthCLIP provides richer input, the fixed CLIP text encoder (especially ViT-B/16) struggles to compress the dense, multi-perspective semantic content into a single discriminative feature vector for T2I tasks. 

\begin{table*}[h]
    \centering
    \small
    \setlength{\tabcolsep}{8pt}
    \begin{tabular}{cccccccccc}
    \toprule
    & \multirow{2}{*}{Method}  & \multirow{2}{*}{Data} & \multicolumn{2}{c}{MS-COCO} & \multicolumn{2}{c}{Flickr-1k} & \multicolumn{2}{c}{Urban-1k}& \multirow{2}{*}{Avg}\\
    \multirow{9}{*}{B/16}& & & I-to-T & T-to-I & I-to-T & T-to-I & I-to-T & T-to-I \\
    \midrule
    &CLIP & 400M & 53.1 & 32.7 & 82.4 & 63.5 & 67.2 & 54.3& 58.87\\
    &Fine-tuned CLIP & 1M & 56.6 & 37.3 & 84.4 & 66.8 & 91.7 & 90.5& 71.22\\
    &Long-CLIP& 1M & 57.6 & 40.7 & 86.9 & 73.7 & 79.0 & 79.3 & 69.53\\
    & FIX-CLIP& 1M & 60.9 & \underline{44.8} & 88.4 &\underline{77.4} & 80.9 &81.1 & 72.25 \\
    & FIX-CLIP & 5M & \underline{61.3} & \textbf{47.0} & \underline{89.1} 
    &\textbf{79.5} & 88.0 &90.8 & \underline{75.95} \\
    & SynthCLIP & 10M & 50.0 & 36.7 & 81.8 & 64.0 & 70.0 & 61.2 & 60.62\\
    & SynthCLIP & 20M & 57.8 & 42.5 & 87.5 & 69.2 & 73.1 & 63.1 & 65.53\\
    \cline{2-10}

    &SigLIP& 12B & 37.3 & 34.5 & 75.2 & 76.4 & 78.2 & 79.6& 63.53\\
    &LoTLIP& 100M & 59.7 & 38.1 & 86.9 & 65.2 & 77.8 & 76.5& 67.36\\
    &Role-SynthCLIP & 1M & \textbf{64.1} & 43.2 & \textbf{89.3} & 73.3 & \textbf{96.3} & \textbf{95.9}&\textbf{77.01}\\
    
    \midrule
    \multirow{4}{*}{L/14}&CLIP& 400M & 56.9 & 35.1 & 86.3 & 64.8 & 68.7 & 56.0 &61.30\\
    &Fine-tuned CLIP & 1M & 66.3 & 47.1 & 91.1 & 77.3 & 94.8 & 95.1& 78.62\\
    &Long-CLIP & 1M & 62.8 & 46.3 & 90.9 & \textbf{78.7} & 86.2 & 82.6& 74.58\\
    &Role-SynthCLIP & 1M & \textbf{68.6} & \textbf{48.8} & \textbf{92.1} & \underline{78.0} & \textbf{97.3} & \textbf{97.8}&\textbf{80.43}\\
    \bottomrule
    \end{tabular}
    \caption{Zero-shot image-text retrieval (Recall@1). Results on MS-COCO, Flickr-1k, and Urban-1k. “Avg” denotes the mean over six metrics (I-to-T and T-to-I on three datasets). Best results are in \textbf{bold}; second best are \underline{underlined}.}
    \label{tab:main_result}
\end{table*}

\noindent \textbf{Zero-shot Classification.}
For the CLIP-B/16 model, Role-SynthCLIP achieves average accuracy (69.62\%) nearly matching the performance of the original CLIP (70.30\%). Importantly, our method exhibits superior performance on Out-of-Distribution (OOD) datasets, achieving the highest accuracy on ImageNet-O (44.5\%) and competitive results on CIFAR-10/100. This suggests that the multi-perspective captions, by capturing diverse attributes beyond simple category names, successfully train a more robust and generalizable feature space capable of handling domain shifts.

In the CLIP-L/14 setting, Role-SynthCLIP attains an average accuracy of 73.56\%, slightly below the original CLIP (73.96\%). Notably, it achieves the best accuracy on ImageNet-O (36.4\%), suggesting that our data augmentation contributes positively to robustness and semantic diversity for challenging OOD tasks, even without yielding the overall highest accuracy.

\begin{table}[]
    \centering
    \small
    \setlength{\tabcolsep}{3pt}
    
    \begin{tabular}{ccccccccc}
    \toprule
    &Models  &  \begin{sideways}IN-1k\end{sideways} & \begin{sideways}IN-o\end{sideways} &\begin{sideways}CIFAR-10\end{sideways}& \begin{sideways}CIFAR-100\end{sideways}& \begin{sideways}Food-101\end{sideways}& \begin{sideways}Avg\end{sideways}\\
    \midrule
    \multirow{4}{*}{B/16}&CLIP & \textbf{68.3} & 40.4 &  90.7 & 66.7 & \textbf{85.4} & \textbf{70.30}\\
    & SynthCLIP-20M & 30.9 & 17.1 & 87.7 & \textit{68.5} & 75.3 &  55.90 \\
    &Fine-Tuned CLIP & 55.1 & 31.7  & 83.9 & 59.2 & 75.3 & 61.04\\
    &Long-CLIP & \underline{66.8} & \underline{42.7}  &  \underline{90.7} & \textbf{69.3} & 81.9 & \underline{70.28}\\
    &Role-SynthCLIP  & 64.8 & \textbf{44.5} &\textbf{91.1} & \underline{68.2} & \underline{79.5} & 69.62\\
    \midrule
    \multirow{4}{*}{L/14}&CLIP & \textbf{75.5} & 31.6 &  \textbf{95.8} & 76.0 & \textbf{90.9} & \textbf{73.96}\\
    &Fine-Tuned CLIP &71.4 & 35.1 &95.2 & 75.6 & 87.0 & 72.86\\
    &Long-CLIP & \underline{75.5} & \underline{33.7} &  \underline{95.3} & \textbf{78.5} & 86.4 & 73.88\\
    &Role-SynthCLIP  & 71.9 & \textbf{36.4} & 95.8 & \underline{77.3} & \underline{86.4} & 73.56\\
    \bottomrule
         \end{tabular}
    \caption{Top-1 accuracy results of zero-shot image classification in the above five validation sets. Best result is in \textbf{bold}; second best are \underline{underlined}.}
    \label{tab:cls}
\end{table}

\subsection{Ablation Study}
\label{sec:ablation}
\noindent \textbf{Effects of Different Experts.}
In our experimental design, various experts evaluated the same image from different perspectives using a role-play approach. To determine the contribution of each expert, we conducted an ablation study on each one. The results are shown in Table \ref{tab:abla_experts}.

Among individual roles, removing the observer or composition analyst leads to the largest drops in accuracy, indicating their importance in grounding captions to visual structure and contextual coherence. The interpreter and narrative setter also provide measurable gains by enriching semantic diversity. These results suggest that each expert captures a complementary aspect of visual understanding, and that their joint conditioning is essential for robust cross-modal alignment.
\begin{table}[htp]
    \small
    \setlength{\tabcolsep}{3pt}

    \centering
    \begin{tabular}{ccccccc}
    \toprule
    \multirow{2}{*}{Method} &  \multicolumn{2}{c}{COCO} & \multicolumn{2}{c}{Flickr}&\multicolumn{2}{c}{Urban}\\
    & I-to-T & T-to-I &I-to-T &T-to-I&I-to-T &T-to-I\\
    \midrule
    Role-SynthCLIP  &  \textbf{68.6} & \textbf{48.8} & \underline{92.1}& \underline{78.0} & \textbf{97.3} &\textbf{97.8}\\
    w/o interpreter  &  67.0 & \underline{48.2} & 92.1& \textbf{78.4} & 96.9 &\underline{97.7}\\
    w/o emotion  &  \underline{67.7} & 48.0 & \textbf{92.3}& 78.4 & 96.9 &97.6\\
    w/o com analyst  &  67.4 & 47.4 & 91.9& 77.7 & 96.4 &96.7\\
    w/o narrative setter  &  67.6 & 47.5 & 92.0& 78.2 & 96.9 &97.6\\
    w/o observer  &  66.7 & 47.2 & 90.9& 77.5 & 96.9 &97.0\\
    \bottomrule
    \end{tabular}
    \caption{Experimental results upon removing each individual expert.}
    \label{tab:abla_experts}
\end{table}

\noindent \textbf{Impact of Caption Generation Strategies.}
The generation of high-quality short captions is crucial for traditional CLIP training, which often prioritizes core, objective semantics. While our primary method utilizes Role-SynthCLIP (Direct Gen.), employing Qwen2.5 VL to generate both long and short captions simultaneously, we conducted an ablation study to compare this strategy against three important baselines based on existing approaches:

\textbf{First Sentence}: Following the approach of Long-CLIP \cite{longclip}, we used the long captions from the original ShareGPT4V dataset and extracted their first sentence to serve as the short caption. This strategy assumes the most critical information is at the beginning.

\textbf{Summarization}: We utilized the long captions from the original ShareGPT4V dataset and employed a Large Language Model (LLM) to perform abstractive summarization, generating the short captions. This tests the benefit of abstractive information distillation.

\textbf{Random Extract}: A sentence was randomly selected from the long caption to serve as the short caption. This acts as a simple control to quantify the contribution of specific extractive strategies.

\textbf{Multi Prompts}: Referring to FIX-CLIP \cite{wang2025fixclipdualbranchhierarchicalcontrastive}, use multiple short prompts, and randomly select a prompt during generation to create the caption.

The results of this ablation (detailed in Table \ref{tab:prompts_strategy}) validate that the Role-SynthCLIP (Direct Gen.) approach, by guiding the MLLM to generate concise, perspective-aware captions from scratch, yields superior performance compared to post-processing or reliance on extractive baselines.

\begin{table}[htp]
    \setlength{\tabcolsep}{3pt}
    \small
    \centering
    \begin{tabular}{ccccccc}
    \toprule
    \multirow{2}{*}{Method} &  \multicolumn{2}{c}{COCO} & \multicolumn{2}{c}{Flickr}&\multicolumn{2}{c}{Urban}\\
    & I-to-T & T-to-I &I-to-T &T-to-I&I-to-T &T-to-I\\
    \midrule
    Role-SynthCLIP  &  \textbf{68.6} & \textbf{48.8} & \textbf{92.1}& \textbf{78.0} & \textbf{97.3} &\textbf{97.8}\\
    First Sentence & 58.7 & 40.2 & 88.9& 77.0 & 78.4 &79.1\\
    Summarization & 60.1 & 41.2 & \underline{89.9}& \underline{78.0} & 88.4 & 89.1\\
    Random Extract & 56.1 & 34.2 & 81.3& 76.1 & 75.2 & 86.0\\
    Multi Prompts  &  \underline{64.3} & \underline{45.2} & 89.6& 75.0 & \underline{95.8} &\underline{96.4}\\
    \bottomrule
    \end{tabular}
    \caption{Comparison of Diverse Caption Generation Methods on Retrieval Tasks. We evaluate our novel Role-SynthCLIP (direct generation of concise, perspective-aware captions) against established approaches, including post-processing methods (Extractive, Abstractive, Random Extract) and the multi-prompts strategy.}
    \label{tab:prompts_strategy}
\end{table}

Our Role-SynthCLIP (Direct Gen.) method consistently and significantly outperforms all investigated baselines across all metrics. For the challenging MS-COCO retrieval task, our strategy achieves 68.6\% I-to-T and 48.8\% T-to-I, establishing a substantial lead over the next best alternative, Multi Prompts (64.3\% I-to-T, 45.2\% T-to-I).

This performance gap validates a crucial insight: relying on post-processing methods, whether extractive (First Sentence, Random Extract) or abstractive (Summarization) will introduce measurable semantic noise and alignment ambiguity.

\noindent \textbf{Impact of Image Pair Filter.}
We eliminated the filtering process and, for comparison, instead randomly sampled a portion of the captions generated by the various experts, specifically 1/number of experts. The results validated the effectiveness of our filtering method.
\begin{table}[htp]
    
    \setlength{\tabcolsep}{3pt}
    \small
    \centering
    \begin{tabular}{ccccccc}
    \toprule
    \multirow{2}{*}{Method} &  \multicolumn{2}{c}{COCO} & \multicolumn{2}{c}{Flickr}&\multicolumn{2}{c}{Urban}\\
    & I-to-T & T-to-I &I-to-T &T-to-I&I-to-T &T-to-I\\
    \midrule
    Role-SynthCLIP  &  \textbf{68.6} & \textbf{48.8} & \textbf{92.1}& \textbf{78.0} & \textbf{97.3} &\textbf{97.8}\\
    w/o filter  &  67.0 & 48.2 & 90.1& 77.4 & 96.1 &96.4\\
    \bottomrule
    \end{tabular}
    \caption{Ablation study on the \textbf{Role-Aware Filter}. We compare the performance of our full model against a setting (\textit{w/o filter}) where generated captions are randomly subsampled at a rate of $1/N_{\text{experts}}$ to match the final data size, demonstrating the filter's necessity in eliminating noisy and role-inconsistent pairs.}
    \label{tab:placeholder}
\end{table}
\subsection{Text-to-Image Generation}
\begin{figure}
    \centering
    \includegraphics[width=\linewidth]{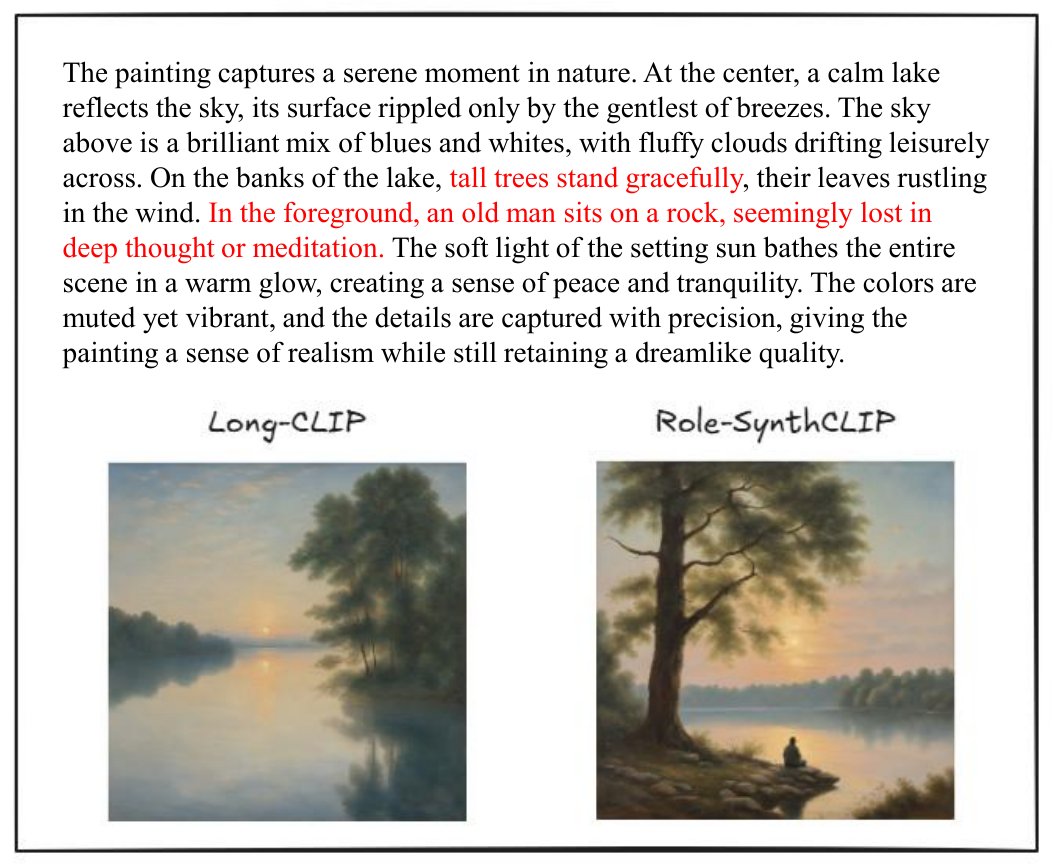}
    \caption{Comparison of text-to-image generation using Long-CLIP and Role-SynthCLIP as text encoders. The prompt describes ``a serene lakeside scene with tall trees and an old man sitting on a rock.'' Role-SynthCLIP better preserves these semantic details, accurately depicting both the human subject and the tall trees.}
    \label{fig:sd}
\end{figure}
Unlike prior improvements that modify CLIP's architecture, we did not alter the structure of the CLIP model but instead preserved its original architecture. This approach allows models such as Stable Diffusion XL to inherit stronger text understanding capabilities, enabling them to generate images that accurately match the input, all without requiring extensive modifications.

As shown in Figure~\ref{fig:sd}, when using Long-CLIP as the text encoder, the generated image fails to capture key semantic elements, omitting the old man and producing undersized trees. In contrast, Role-SynthCLIP yields a faithful depiction consistent with the input prompt, demonstrating improved semantic alignment and compositional grounding in text-to-image generation.

\section{Discussion}
We aim to address two key questions that impact the experimental outcomes: (i) how role-playing affects the caption sampling results, and (ii) how our method influences the model's performance.
\subsection{How role-play influences the experts' behavior}
Our analysis reveals that role play fundamentally restructures how multimodal models allocate attention across the visual and linguistic modalities. Using token-level attention visualization (TAM) \cite{li2025tokenactivationmapvisually} to probe the inner mechanisms of the multimodal large language model (MLLM), we observe a systematic shift in image-oriented attention patterns depending on the semantic abstraction level of the assigned role.
\begin{figure}
    \centering
    \includegraphics[width=\linewidth]{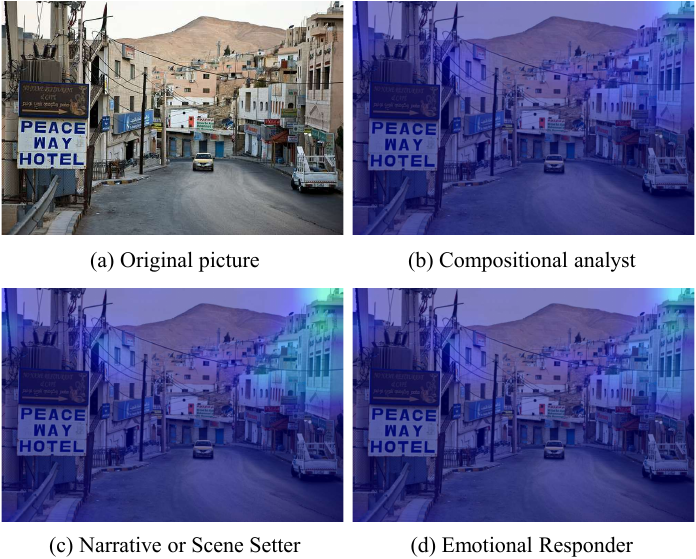}
    \caption{Role-conditioned attention behavior in Role-SynthCLIP. Token Activation Map (TAM) visualizations compare caption token ``buildings'' produced under three expert roles Narrative or Scene Setter, Emotional Responder, and Compositional Analyst-against the original image.}
    \label{fig:dis_1}
\end{figure}

Our analysis of token-level attention patterns reveals intriguing insights into how different expert roles in Role-SynthCLIP influence attention distribution across image and text modalities. Specifically, we observe that roles focused on abstract concepts, such as the Narrative or Scene Setter and emotional perspectives (e.g., Emotional Responder), allocate significantly more attention to the image compared to roles that focus on lower-level semantic elements, such as compositional analysis.

As illustrated in Figure \ref{fig:dis_1}, the Narrative Setter, tasked with generating captions that embody storytelling elements, demonstrates a notable increase in attention to visual cues that relate to the overall structure or emotional context of the scene. This tendency aligns with the need for the model to integrate multiple aspects of the visual content in a more abstract, global manner. On the other hand, roles focused on more compositional aspects, such as the Compositional Analyst, display a more restrained attention allocation to the visual domain, even when using similar or identical words. Its task is not to describe the buildings' appearance, but their spatial relationship to other elements. This focus on relational structure, rather than holistic context, appears to demand a more localized and specific attentional focus.

This pattern suggests that attention distribution is closely tied to the cognitive task assigned to each expert role. Abstract, higher-level concepts often require a broader, more integrated view of the scene, which leads to increased attention across the image. In contrast, tasks such as compositional analysis, which involve focusing on specific visual structures or smaller image components, result in more localized attention.

These findings underscore the importance of role-playing in steering the model's attention toward different levels of semantic richness. The ability to simulate multiple expert perspectives, each emphasizing distinct aspects of the visual content, not only enhances caption diversity but also improves the model's understanding of complex visual phenomena.

\subsection{Qualitative Visualization Analysis between CLIP and Role-SynthCLIP}
\begin{figure}
    \centering
    \includegraphics[width=\linewidth]{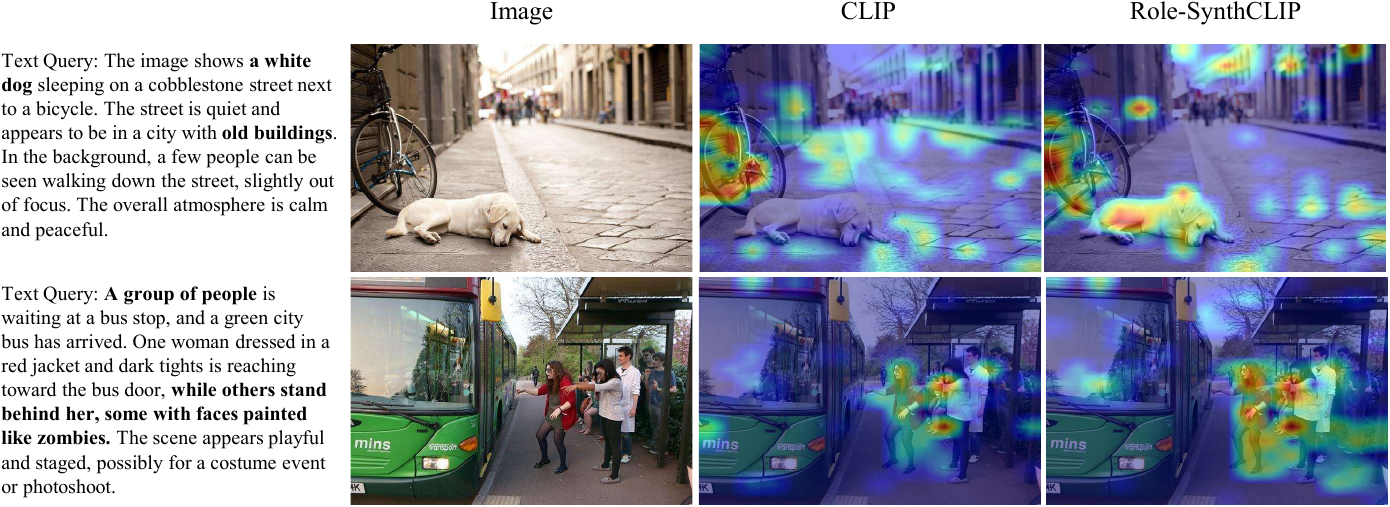}
    \caption{Qualitative comparison of attention maps between CLIP and Role-SynthCLIP. We visualize gradient-based saliency (Grad-CAM) from the text \texttt{[CLS]} token to image patch tokens to compare CLIP and Role-SynthCLIP.}
    \label{fig:dis_2}
\end{figure}
We further compare attention distributions between Role-SynthCLIP and the original CLIP to understand how role play affects visual grounding. As shown in Figure \ref{fig:dis_2}, Role-SynthCLIP consistently exhibits more semantically aligned and context-sensitive attention patterns.

In the first case (a white dog sleeping on a cobblestone street), our model focuses sharply on the foreground subject and architectural contours along the street, while CLIP disperses its attention over irrelevant background regions. This indicates that the multi-role synthesis effectively guides the model to prioritize visually and semantically coherent cues.

In the second case (a group of people at a bus stop), Role-SynthCLIP shifts its focus toward the background crowd and human interactions, capturing the social and narrative context that CLIP largely overlooks. The model demonstrates an enhanced ability to integrate relational semantics and global scene understanding rather than fixating on isolated objects.

Overall, these observations suggest that role-driven data generation restructures the model’s attention hierarchy. Instead of static saliency, Role-SynthCLIP dynamically allocates attention according to the underlying cognitive intent implied by each role, balancing detailed perception with contextual reasoning. This adaptability underscores the key advantage of role play in fostering deeper and more flexible vision–language alignment.

\section{Conclusion}
We introduced Role-SynthCLIP, a novel framework that leverages role-playing prompts to generate high-quality synthetic data for vision-language models. By simulating diverse expert perspectives, our method enhances the semantic diversity and quality of captions, addressing common issues of data noise and redundancy in large-scale datasets. Experimental results demonstrate that Role-SynthCLIP achieves state-of-the-art performance with only 1M synthetic pairs, surpassing models trained on significantly larger datasets. This work provides a data-efficient solution for training vision-language models, offering new avenues for controlled data generation and advancing the performance of multimodal tasks.

{
    \small
    \bibliographystyle{ieee_fullname_unsrt}
    \bibliography{main}
}

\clearpage
\setcounter{page}{1}
\maketitlesupplementary

\section{Expert Role Generation Prompts}
\label{sec:appendix_expert_role_prompts}
Inspired by LLM-Discussion \cite{lu2024llmdiscussionenhancingcreativity}, we use GPT-5 to automatically generate the roles used in our framework. These roles are not intended to be optimal; rather, they are preliminary designs to demonstrate the effectiveness of our approach. We also acknowledge that defining a broader spectrum of roles would further enhance data diversity, which we will explore in future work. The detailed prompt is shown in Table \ref{tab:suppl_1}.
\begin{table*}[!ht]
    \centering
        \small
    \begin{tabular}{p{0.95\linewidth}}
    \toprule
    Round 1 - User Prompt \\
    \midrule
     What are some roles that are typically used in boosting precise image description, like different personas to precisely describe the image in different view.    \\
    \bottomrule
\end{tabular}
\end{table*}

\begin{table*}[!ht]
    \centering
        \small
    \begin{tabular}{p{0.95\linewidth}}
    \toprule
    Round 1 - GPT-5 Response \\
    
    \midrule
     That’s a really interesting question! When people try to boost the precision of image descriptions, they often use different roles or personas to look at the picture from complementary perspectives. This technique is used in annotation, accessibility work, and in multimodal AI prompting to generate richer and more precise outputs.    \\

     Here are some typical roles that can help:

1. The Observer of Details
Focuses on the specific visual attributes of an image such as objects, colors, textures, shapes, and lighting conditions. Strong emphasis on capturing factual, observable details.

2. Interpreter of Context
Identifies the possible meaning of visual elements, interpreting human expressions, implied actions, cultural references, and the situational context of the image.

3. Compositional Analyst
Examines the overall structure of the image, including arrangement of subjects, balance, perspective, depth, and spatial relationships. Highlights the visual organization and framing.

4. Narrative or Scene Setter
Synthesizes observed details and context to suggest a narrative, identifying the setting, possible actions, implied storylines, and the overall communicative intent of the image.

5. Emotional or Aesthetic Responder
Focuses on the image’s emotional resonance and aesthetic qualities, capturing mood, atmosphere, symbolic associations, and the subjective impression it may evoke in viewers.

The idea is that by combining multiple roles, you get a layered and more accurate description—moving from raw perception to interpretation, while recognizing where assumptions end and evidence begins. \\
    
        \bottomrule
\end{tabular}
\end{table*}

\begin{table*}[!ht]
    \centering
        \small
    \begin{tabular}{p{0.95\linewidth}}
    \toprule
    Round 2 - User Prompt \\
    \midrule
     Great, now follow this format and generate a .json file for these roles: Remain ”type”: ”openai”, ”model name”: ”gpt-3.5-turbo”, the same. and the agent name = ”GPT
Agent 1 - {agent role}”

\begin{lstlisting}[language=json,firstnumber=1]
[
  {
    "type": "openai",
    "model_name": "gpt-3.5-turbo",
    "agent_name": "GPT Agent 1 - White Hat",
    "agent_role": "White Hat",
    "agent_speciality": "Information Analysis and Facts",
    "agent_role_prompt": "Focuses on available data and past information, analyzing trends and gaps in knowledge, striving for an objective viewpoint."
  },
  {
    "type": "openai",
    "model_name": "gpt-3.5-turbo",
    "agent_name": "GPT Agent 2 - Red Hat",
    "agent_role": "Red Hat",
    "agent_speciality": "Emotions and Feelings Interpretation",
    "agent_role_prompt": "Listens to and validates the emotional responses of the group, understanding the values and intuition behind reactions, without judgment or justification."
  }
]
\end{lstlisting}    \\
            \bottomrule
\end{tabular}

\end{table*}
\begin{table*}[!ht]
    \centering
        \small
    \begin{tabular}{p{0.95\linewidth}}
    \toprule
    Round 2 - GPT-5 Response \\
    \midrule
     Below is the .json format including all the personas you specified:
     
\begin{lstlisting}[language=json,firstnumber=1]
[
  {
    "type": "openai",
    "model_name": "gpt-3.5-turbo",
    "agent_name": "GPT Agent 1 - Observer of Details",
    "agent_role": "Observer of Details",
    "agent_speciality": "Micro-level visual recognition",
    "agent_role_prompt": "Focuses on the specific visual attributes of an image such as objects, colors, textures, shapes, and lighting conditions. Strong emphasis on capturing factual, observable details."
  },
  {
    "type": "openai",
    "model_name": "gpt-3.5-turbo",
    "agent_name": "GPT Agent 2 - Interpreter of Context",
    "agent_role": "Interpreter of Context",
    "agent_speciality": "Semantic and situational interpretation",
    "agent_role_prompt": "Identifies the possible meaning of visual elements, interpreting human expressions, implied actions, cultural references, and the situational context of the image."
  },
  {
    "type": "openai",
    "model_name": "gpt-3.5-turbo",
    "agent_name": "GPT Agent 3 - Compositional Analyst",
    "agent_role": "Compositional Analyst",
    "agent_speciality": "Macro-level composition analysis",
    "agent_role_prompt": "Examines the overall structure of the image, including arrangement of subjects, balance, perspective, depth, and spatial relationships. Highlights the visual organization and framing."
  },
  {
    "type": "openai",
    "model_name": "gpt-3.5-turbo",
    "agent_name": "GPT Agent 4 - Narrative Setter",
    "agent_role": "Narrative or Scene Setter",
    "agent_speciality": "Story and situational framing",
    "agent_role_prompt": "Synthesizes observed details and context to suggest a narrative, identifying the setting, possible actions, implied storylines, and the overall communicative intent of the image."
  },
  {
    "type": "openai",
    "model_name": "gpt-3.5-turbo",
    "agent_name": "GPT Agent 5 - Emotional Responder",
    "agent_role": "Emotional/Aesthetic Responder",
    "agent_speciality": "Mood, tone, and subjective experience",
    "agent_role_prompt": "Focuses on the image's emotional resonance and aesthetic qualities, capturing mood, atmosphere, symbolic associations, and the subjective impression it may evoke in viewers."
  }
]
\end{lstlisting} \\
\bottomrule 
\end{tabular}
\caption{Prompts used for role generation.}
\label{tab:suppl_1}
\end{table*}

\section{Expert Captioner Observation Prompts}
We use the \textit{agent role}, \textit{agent speciality}, and \textit{agent role prompt} to define distinct character profiles, enabling the MLLM to generate captions aligned with its assigned role. The detailed prompt is presented in Table~\ref{tab:suppl_2}.

\label{sec:appendix_expert_captioner_prompts}
\begin{table}[ht]
    \centering
        \small
    \begin{tabular}{p{0.95\linewidth}}
    \toprule
    Prompt for VLM to generate captions \\
    \midrule
     You are a \{agent role\} whose specialty is \{agent speciality\}. \{agent role prompt\}.\\ 
You will create an descriptions (max 150 words)/(max 30 words for short caption) for the given image, from the perspective of an \{agent role\}, highlighting \{agent speciality\}. \\
Output only the descriptions with no extra explanations. Do NOT use any title words such as: "Mood:","Tune:","Joyful simplicity:","Muted Precision:","Elegant simplicity:","Muted Precision: "    \\
    \bottomrule 
    \end{tabular}
    \caption{Prompt used for caption generation.}
    \label{tab:suppl_2}
\end{table}

The sampling parameters we use are show in Table~\ref{tab:generation_params}. The parameters are configured following the official recommendations of Qwen.

\begin{table}[ht]
\centering
    \small
\begin{tabular}{lc}
\toprule
\textbf{Parameter} & \textbf{Value} \\
\midrule
top\_k & 1 \\
top\_p & 0.001 \\
temperature & 0.01 \\
repetition\_penalty & 1.0 \\
presence\_penalty & 1.5 \\
frequency\_penalty & 0.0 \\
\bottomrule
\end{tabular}
\caption{Hyperparameter settings for generation.}
\label{tab:generation_params}
\end{table}

\section{Text Image Pair Filter Prompt}
Captions generated by multimodal large language models (MLLMs) may contain two types of errors:  
(i) inconsistencies between image and text, and  
(ii) deviation from the assigned role.  

To address this, we leverage the teacher model GPT-5 to obtain training data containing both scores and reasoning paths. Using this data, we train a smaller and more cost-efficient filtering model to score image-text pairs. The filtering prompts used are shown in Table~\ref{tab:suppl_3}. 
\label{sec:appendix_filter_prompt}

\begin{table}[!ht]
    \small
    \centering
    \begin{tabular}{p{0.95\linewidth}}
    \toprule
    Prompt for VLM to score an image-text pair \\
    
    \midrule
     Here is Text Caption from a \{agent role\}, which role is \{agent role prompt\}\\
Text Caption: \{caption\}\\

Please evaluate if the provided text caption accurately represents the main features and objects of the image. The caption doesn't need to detail every aspect of the image, but it should capture its primary theme. Rate the overall quality of the text caption's match to the image on a scale of 1-100, considering the criteria mentioned. A higher score indicates higher level of image text matching. \\

Ensure that your scoring is nuanced and uses the entire range from 0 to 100, reflecting the subtle differences. The score should be given as an integer, with each number between 0 and 100 considered as a potential score, avoiding the tendency to round to multiples of 5 or 10. Please first output a single line containing the value indicating the scores. You must consider the role of the captioner when evaluating the caption. In the subsequent line, please provide a comprehensive explanation of your evaluation, avoiding any potential bias.   \\
    \bottomrule 
    \end{tabular}
    \caption{Prompt for scoring image–text alignment and role consistency.}
    \label{tab:suppl_3}
\end{table}

\section{Filter Algorithm}
\label{sec:appendix_filter_algorithm}
Following SynthCLIP~\cite{hammoud2024synthclipreadyfullysynthetic}, image diversity is a key factor influencing model performance. While maintaining the quality of image–text pairs, we prioritize maximizing image diversity. After filtering, our dataset contains 879,406 unique images, each associated with an average of 3.81 distinct captions, reflecting both visual richness and linguistic variability. The pseudocode of our filtering algorithm is provided in Algorithm~\ref{alg:role_filter}. 

\begin{algorithm}[htp]
    \small
\caption{Role-Aware Cap-and-Refill Filtering Algorithm}
\label{alg:role_filter}
\begin{algorithmic}[1]

\Require Dataset $D$ with fields (\texttt{image}, \texttt{caption}, \texttt{role}, \texttt{score})
\Require Hyperparameters $K_{\max}$, $K_{\min}$, \texttt{TARGET\_PAIRS}

\Statex
\State \textbf{Prefilter:} For each role $r$, keep the top $50\%$ samples ranked by \texttt{score}.
\If{\texttt{dedup} = True}
    \State Remove duplicates based on (\texttt{image}, \texttt{caption}).
\EndIf

\Statex
\State \textbf{Role-wise Normalization:}
\State $s_{i} \leftarrow \dfrac{x_{i}-\mu_r}{\sigma_r + \varepsilon}$ \text{ for each role } $r$

\Statex
\State \textbf{Global Sorting:} Sort all rows by normalized score $s_i$ in descending order.

\Statex
\State \textbf{Initial Selection:} 
\State For each \texttt{image}, keep up to $K_{\max}$ highest-scoring captions.
\State Let $K$ be the current kept set.

\Statex
\If{$|K| > \texttt{TARGET\_PAIRS}$}  \Comment{Over budget — pruning}
    \State Keep at least $K_{\min}$ per \texttt{image}.
    \State Fill remaining quota by descending score order.
    \State \Return $K$
\ElsIf{$|K| < \texttt{TARGET\_PAIRS}$} \Comment{Under budget — refill}
    \State Add top-1 caption for uncovered images.
    \State Refill remaining quota from candidate pool by score, respecting $K_{\max}$.
\EndIf

\Statex
\State \Return Final filtered dataset $K$.

\end{algorithmic}
\end{algorithm}

\section{Training Hyperparameters}
For a fair comparison, we adopt the same training parameters as Long-CLIP. Due to limited computational resources, we apply gradient accumulation to compensate for the smaller GPU count and maintain the same global batch size as Long-CLIP. The detailed training settings are summarized in Table~\ref{sec:training_hypers}.

\begin{table}[!ht]
        \small
    \centering
    \begin{tabular}{lc}
    \toprule
    \textbf{Configuration} & \textbf{Role-SynthCLIP Training} \\
    \midrule
    Global batch size & 2048\\
    Training epochs & 6 \\
    Learning rate & 1e-6 \\
    Warm-up steps & 200 \\
    Learning rate scheduler & cosine \\
    Optimizer & AdamW \cite{loshchilov2019decoupledweightdecayregularization} \\
    Optimizer hyperparameters & $\beta_1, \beta_2, \epsilon = 0.9, 0.999, 1\text{e-}8$ \\
    Weight decay & 1e-2 \\
    \bottomrule

\end{tabular}
    
    \caption{Training hyperparameters for Role-SynthCLIP.}
\end{table}
\label{sec:training_hypers}


\end{document}